# Statistical patterns of word frequency suggesting probabilistic nature of human languages

**Shuiyuan Yu[a], Chunshan Xu[b], Haitao Liu[c,d,a*]**

a. Institute of Quantitative Linguistics, Beijing Language and Culture University, 100083, Beijing, China.

b. School of Foreign Studies, Anhui Jianzhu University, 230601, Hefei, China.

c. Department of Linguistics, Zhejiang University, 310058, Hangzhou, China.

d. Centre for Linguistics and Applied Linguistics, Guangdong University of Foreign Studies, 510420, Guangzhou, China.

*Correspondence to: Haitao Liu, e-mail: lhtzju@gmail.com

**Abstract**

Traditional linguistic theories have largely regard language as a formal system composed of rigid rules. However, their failures in processing real language, the recent successes in statistical natural language processing, and the findings of many psychological experiments have suggested that language may be more a probabilistic system than a formal system, and thus cannot be faithfully modeled with the either/or rules of formal linguistic theory. The present study, based on authentic language data, confirmed that those important linguistic issues, such as linguistic universal, diachronic drift, and language variations can be translated into probability and frequency patterns in parole. These findings suggest that human language may well be probabilistic systems by nature, and that statistical may well make inherent properties of human languages.

## 1. Introduction

Language makes a vital part of human intelligence and cognition, a medium of information communication [1] and culture [2], and a vehicle of human thinking and knowledge [3], which is often believed to differentiate human beings from animals [4, 5]. Hence, to decipher the mystery of human beings and human intelligence, we probably have to understand how language works and evolves. For this purpose, various linguistic theories have been formulated. Since the beginning of 20th century, incessant efforts have been devoted to establishing various formal linguistic theories that try to model human languages in terms of rigid either/or rules [6-8]. However, these theories, though having considerably extended our knowledge of human languages, probably have not sufficiently revealed the secrets of language. For example, these theories often have difficult in accounting for the gradual changes of language. What is more, these theories have largely failed in practical applications like natural language processing [9]. These linguistic theories can be traced back to Saussure[10], the founder of modern linguistics, who drew the distinction between langue [the abstract formal language system] and parole[the authentic language use], contending that the only mission of linguistics is to establish the langue, a formal system of strict rules that represents the essential knowledge of human language. But the wild bushes of authentic language seem to always present so many exceptions to the formal, neat, and clear-cut rules of linguistic theories that these theories often face tremendous difficulties when it comes to process real languages [11]. In contrast, the statistics-oriented approaches based on corpus have reaped stunning achievement in natural language processing [12]. This fact suggests that, probably, those formal theories may oversee a vital property of human language, the frequency and probability in authentic language use, and neglecting this probabilistic nature of human languages may be responsible for the failure of rigid formal linguistic theories to correctly model human language.

So, language is probably more a probabilistic system than a formal system of rigid rule, a probabilistic system grounded in actual linguistic behaviors, consisting not so much in langue as in parole. Linguistic units and their relations at different levels are distributed in probability spaces established through language use, and these probabilities make inherent properties of human languages. This revolutionarily new conception may explain why recent achievements of NLP are largely ascribed to computer scientists and mathematicians, not linguists: these NLP technologies mostly rely on probabilistic models extracted from large data of authentic language use, i.e. the parole, bearing very little on the formal linguistic theories [12]. These successes in NLP strongly suggest that language is probably not, or at least not merely, a formal system, but a probabilistic system well-grounded in authentic language use [13-15].

In fact, some linguistic studies have suggested important roles of frequency and probability in language production, comprehension, and change. Experiments have affirmed frequency and probability as a major factor regulating language comprehension [16-18], influencing language acquisition [19-21], shaping syntactic patterns [22], driving the emergence and evolution of linguistic constructions [23]. Experiments also reported that high frequency words are easy to memorize and recognize [24], that high frequency brings about ready and accurate word judgments [25], and that frequency has much effect on the variation of language in a community [26]. These findings suggest that frequency and probability information of language use are stored in human minds, influencing language comprehension, language production, language acquisition, sociolinguistics, etc.

Language as a probabilistic system rooted in parole is a revolutionarily new idea, which may provide a new perspective for us to understand human cognition, human intelligence, and human languages. But this new idea must be anchored to ample evidences. Cognitive and psychological studies, mostly microscopic and isolated, have presented some empirical evidences. But they are not sufficient: macroscopic and systemic evidences are also indispensable, which is what the present study seeks. If this new idea is right, it may be expected that the important linguistic issues are not qualitative matters of yes/no dichotomies but quantitative matters of probability or frequency distributions. For example, linguistic universals [27-28] may manifest as universal patterns of probability or frequency widely found human languages. In other words, if language is a probabilistic system, some macroscopic statistical patterns might be found in almost all human languages, which are linguistically significant enough to stand as linguistic universals. Similarly, if language is a probabilistic system, the changes of language must be changes in statistical pattern of language use. In this sense, diachronic change[29] and personal styles[30], which are the pivotal concerns of historical linguistics and social linguistics, can also be captured in terms of changes and variations in those overall statistical patterns found in language data. To recap, the statistical patterns of probability or frequency may well have quantifiable changes over time, across regions, and among persons, which may account for diachronic drift, regional varieties, and personal styles. If these fundamental linguistic issues are actually issues of probability and frequency, we may have macroscopic evidences that language is very likely to be a probabilistic system. This type of evidences is sought for the first time in the present study, which conducted the first large-scale statistical investigations into word frequency distribution of multiple languages that spans several hundred years. This study concentrates on word frequency distribution because on one hand, words are probably the least controversial linguistic units, which play fundamental roles in language, and on the other, their frequencies make the basis of virtually all statistical properties found at higher linguistic levels. Our findings suggest that the statistical patterns of word frequency can reflect both linguistic universals and linguistic variations mentioned above, which indicates that statistical patterns may make inherent properties of human languages, and thus supports the idea that language is a probabilistic system [31-32].

## 2. Material and methods

### *2.1. Language materials*

105 languages from 12 language families in Leipzig Corpora [33] are used as language materials for the investigations into word frequency distributions. Google Books Ngram Viewer[34] is used as language materials for the study of diachronic drifts of word frequency distributions. The language materials for explorations into personal variations in word frequency distribution come from Project Gutenberg [35], of which the present study selects 2707 works by 316 authors born in the 19th century. Most of these authors have less than 10 works, with very few of them having more than 10 works. When collecting statistics, we exclude numbers and punctuations, which have little to do with essential properties of language.

### *2.2 Indices of language change*

Cosine similarity and spearman rho are the measures of changes in the kernel lexicon (the set of 3000 most frequent words). Cosine similarity indicates word set similarity, that is, the degree of overlap between two word-sets by measuring the fraction of the intersection size of two sets divided by the geometric mean of the two set sizes, which ranges between 0 and 1. Spearman's rho measures the degree of correlation between 2 word frequency rank orders, that is, the degree of similarity between two word sets in terms of their word frequency rank order.

cosine similarity： $O(A, B) = \frac{|A \cap B|}{\sqrt[2]{|A| \cdot |B|}}$ (1)

Spearman rho： $r_s = 1 - \frac{6 \sum {d_i}^2}{n(n^2-1]}$ (2)

In these two formulae, A and B represent two kernel lexicon, $d_i$ is the order difference between two sequences (two word frequency rank orders), n is the number of observations (sequence length). The value of Spearman rho ranges between - 1 and + 1, and the symbols + and - presents the direction of correlation.

Actually, these two indices are closely related to each other. This study calculates the two indices of six languages in Google Book at time intervals of 1,2,4,8,16,32,64 years, and then calculates the correlation coefficients between these two groups of indices. The results show that the coefficients are very high: for American English, the coefficient is 0.9099, for British English, it is 0.9780, for French it is 0.9858, for German it is 0.9595, for Italian it is 0.9736, for Spanish it is 0.9837.Hence, we can use either of them to measure linguistic similarity.

## *2.3. Personal styles indicated by the kernel lexicon ( the set of the 3000 most frequent words)*

Kernel lexicon may reflect some fundamental language properties. The individual differences in the kernel lexicon hence may mark personal language styles. This study collected, from Gutenberg Project, 2707 works by 316 authors. Word frequency distributions of the 3000 most frequent words (the kernel lexicon) are extracted from all these 2707 works to serve as the standard of comparison. For each word ($w_i$) in these 3000 words, its standard frequency is $f_i$. Thus, we have a frequency set ($f_1, f_2, \ldots\ldots f_{3000}$). Then, we extracted from each work the frequencies of these 3000 most frequent words ($w_i$) , which is another frequency set($q_1$, $q_2$, $\cdots\cdots q_{3000}$). The frequency set found in each work are then compared with this standard frequency set, and the differences between them make a 3000-dimention vector ($q_1$-$f_1$, $q_2$-$f_2$, ……$q_{3000}$-$f_{3000}$) signifying this work. With works of the same author in one group, ANOSIM (Analysis of Similarities) can be carried out to compare differences among authors (between group difference) with differences among works of the same author (in group difference). When ANOSIM yields large R value and small p value, it means that between-group (inter-author) differences are significantly greater than in-group (intra-author) difference, and that vector ($q_1$-$f_1$, $q_2$-$f_2$,……$q_{3000}$-$f_{3000}$) can serve as indicators of personal styles.

The analysis method adopted in this study is Analysis of Similarities (ANOSIM), which, unlike ANOVA (Analysis of Variance), does not require the data value distribution to meet specific conditions.

# 3. Results

## *3.1. Universal patterns in word frequency distribution across human languages*

About 80 years ago, Zipf reported that the frequency of a word in a text or a text-set is a power-law function of its frequency rank order, with an exponent around -1[36]. To verify the universality of Zipf's law, we investigated into the word frequency distributions in the language materials of 105 languages from Leipzig Corpus, with about 1 million words for each language. The results indicate a universal power-law relation between word frequency and frequency rank, as can be seen in Figure 1. This universal pattern of word frequency distribution may be an inherent property of language, reflecting some fundamental mechanisms of language. What is more, the curves of distributions all present a downward bending, which starts, in English, roughly at the 3000th most frequent word, and the curves are thus divided into two parts (Fig. 1), the upper one consisting of frequently used words, mostly function words, referred to as the "kernel lexicon" [37], the lower part, mostly infrequently used words [37-38].

In fact, the word frequency distributions curves in all these 105 languages universally present similar patterns (Fig. 1), with similar linguistic significance. The curves universally consist of two parts. The upper part of the curve covers mostly function words (i.e. segment 1 in Fig.1) and kernel notional words (i.e. segment 2 in Fig.1), which are either the fundamental syntactic means in sentence organization and comprehension or the fundamental blocks in our knowledge of world and semantic representation. The lower part consists of these infrequent content words. This universal pattern of word frequency is probably not a contingent coincidence, but a linguistic universal shaped by fundamental mechanisms like human cognition.

Random languages (monkey typing models) can also present power law distributions of word frequency [39], which, however, differ from those of natural languages in that the power law distributions in random languages usually lack downward bending and present different word length distributions [40].

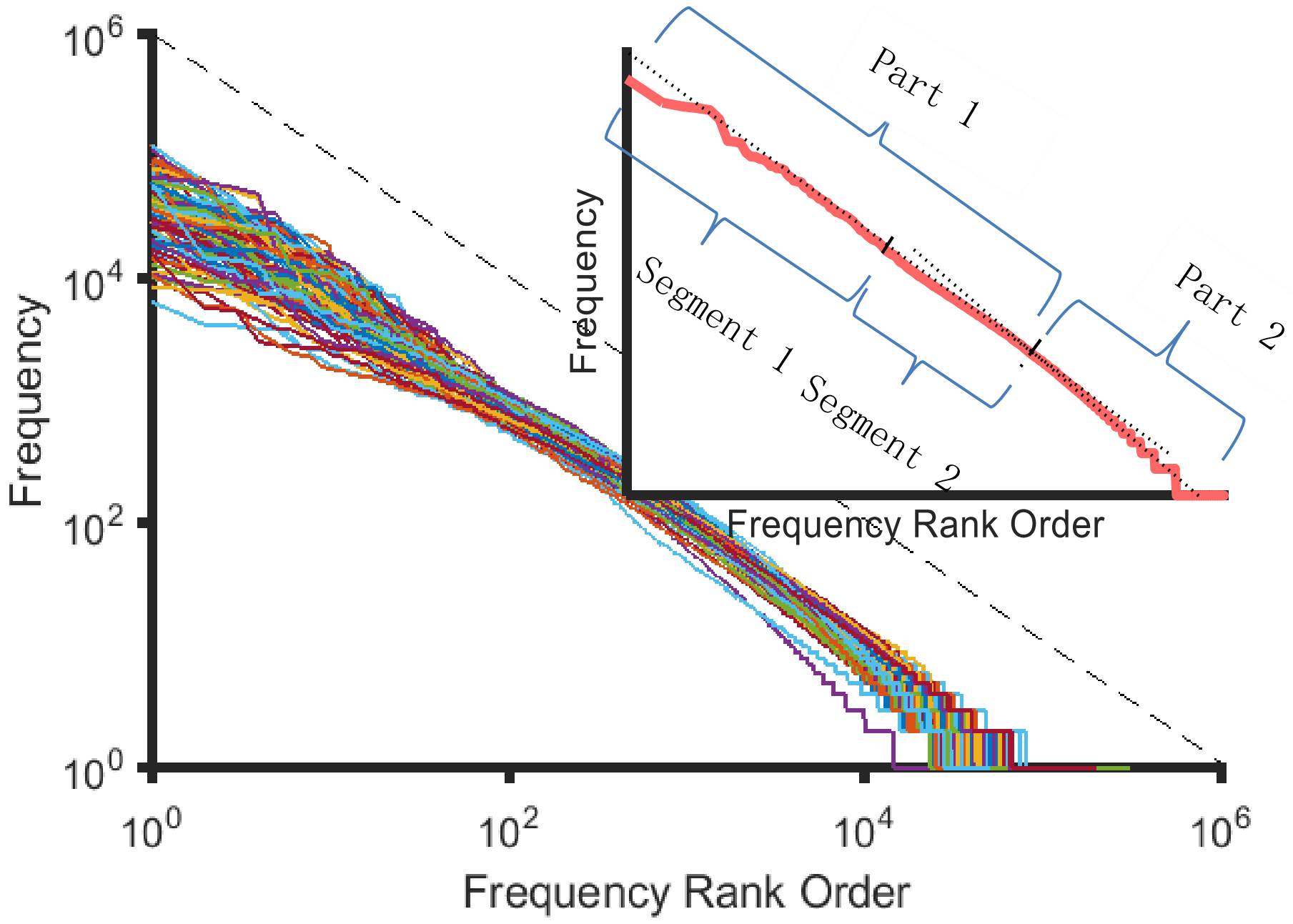

Fig. 1. Word frequency distributions of 105 languages. For these 105 languages from over 10 language families, the word frequency invariably present similar power law distributions, with the downward bending appearing at roughly similar positions on the curves, In fact, the upper part seems to consist of 2 segments[37]. The first one at the top is largely function words, and the second segment below it, mostly the highly frequent content words generally used across different topics. That is, most words in the upper part are function words and kernel notional words, which are probably more inherent and fundamental in the language, reflecting the mutually shared "mean"—the core—of that language, while the words in the lower part are more peripheral, very often motivated by specific topics.

### *3.2. Word Frequency distributions as Indicators of diachronic drift*

Language is always changing. The changes will accumulate over time and eventually lead to significant diachronic differences. Similarly, for language variants in different geographical locations, their differences should also increase over time. If language is a probabilistic system, the gradual diachronic drift should manifest itself in the probability or frequency patterns extracted from language use, that is, the parole. Hence, the present study hypothesized word frequency distributions, which play fundamental roles in language use, may manifest gradual change over time, making an important part of diachronic drift of language.

As can be seen in Figure 1, the curve of word frequency distribution can be roughly divides into two parts at the 3000th most frequent word. In the upper part are those high-frequency words, which make the core of vocabulary, stably used for almost all topics, and in the lower part are the low-frequency words, which are somewhat peripheral, biased toward some specific topics [37]. That is, words in the upper part are probably more inherent and fundamental in the language, reflecting the mutually shared core of that language, while the words in the lower part are more contingent, very often motivated by specific topics. As a result, in comparison with the low-frequency words, the high-frequency core words making the upper part are more suitable for the study of diachronic drift of language because these words and their statistical patterns may reflect the properties of languages.

In this study, the changes in languages are appraised with two indices which measure the changes in this kernel lexicon of 3000 words. The first index is Cosine similarity [41], which measures word set similarity, indicating the degree of similarity (or difference) between two word-sets. Another index is Spearman's rho [41], indicating the degree of similarity (or difference) between two word sets in their word frequency rank order, that is, the degree of similarity in how the words are ordered in terms of their frequencies.

The language materials are taken from Google Book Corpora (https://books.google.com/ngrams), which have about 8 million books published from 1500 to 2009 [42]. For the present study, we select 6 languages in the corpora: American English, British English, French, German, Italian, and Spanish. From the Google n-gram corpora, these two indices of set similarity are extracted to measure the diachronic changes.

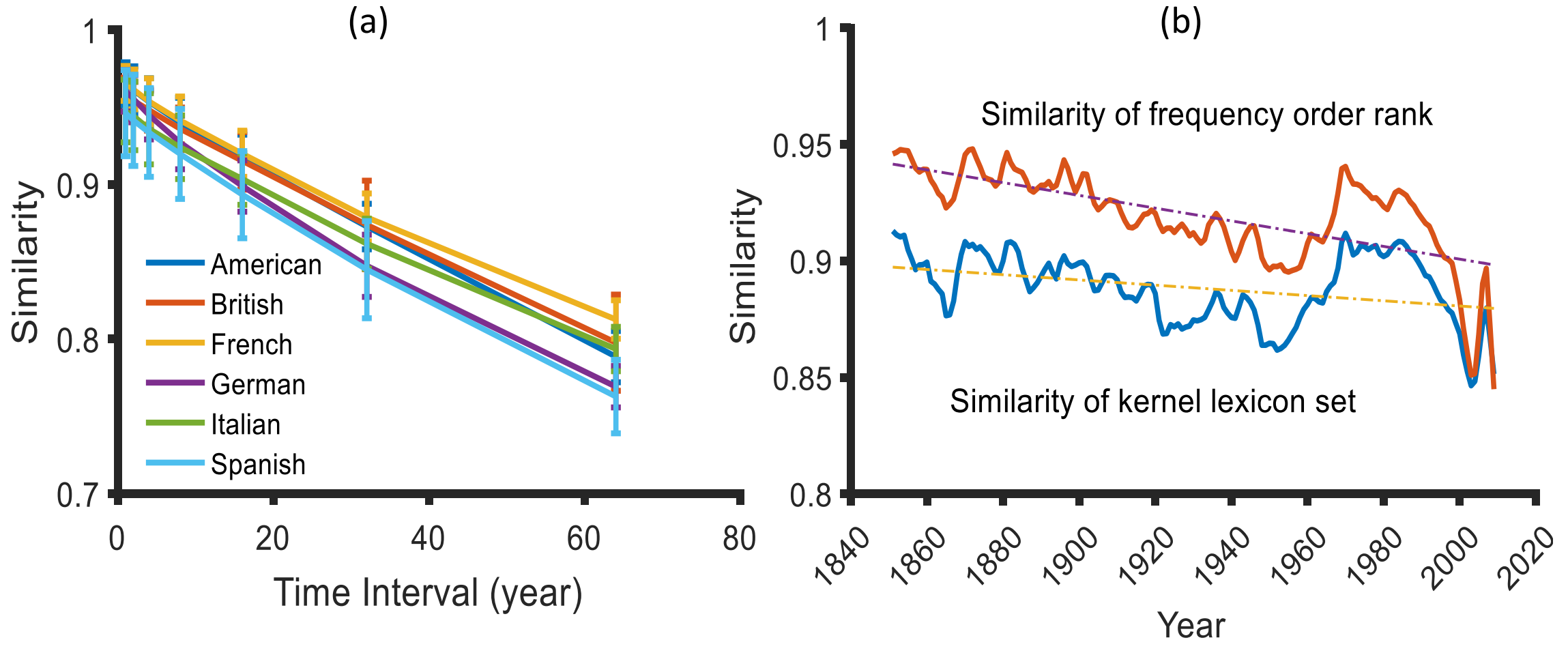

Fig. 2. The diachronic changes of the 3000 most frequent words in 6 languages(a), and the diachronic decrease of similarity between British and American English in 150 years(b) (The curves in (b) are smoothed, using a moving average filter with a span of 5). Figure 2a shows gradual diachronic decrease in similarity (cosine similarity) of 3000 high frequency words in six languages over 150 years, at different time intervals. Figure 2b shows the rather gradual diachronic decrease (spanning 150 years) of similarity between American English and British English in terms of the 3000 most frequent words.

The statistical study of 6 languages suggests that the similarity in the kernel lexicon, though stably high during 150 years, gradually drops over time (Fig.2a). To recapitulate, the 3000 most frequent words in each language are relatively stable, with only minor changes, but the changes accumulate over time. The similarity between the two varieties of English in their 3000 most frequent words also tends to decrease slowly (Fig. 2b). These findings suggest that the statistical patterns of word frequency can well reflect diachronic and regional variations of languages, probably making fundamental properties of language.

### *3.3. Personal styles as reflected by word frequency distributions*

There are, apart from diachronic drifts, individual variations in languages, or, different personal styles. If language is a probabilistic system, the personal styles should register themselves in patterns of probability or frequency. In the present study, it may be expected that such individual differences may be found in the frequency distribution of the kernel lexicon. According to Saussure, the autonomous language system is an idealized "mean" of the miscellaneous use of language, and this system is stored in individuals as slightly different copies. The word frequency distribution extracted from the whole speech community is probably a "mean", and the word frequency distribution stored in an individual, which derive from his or her language experience, is a copy of that mean, that is, the overall frequency distribution. But this copy slightly deviates from the mean and differs from one another, and these differences and deviations may mark personal stylistic features. In other words, the linguistic style of a person may lie partly in the differences between his word frequency distribution and that of the entire speech community. To test this possibility, this study collected, from Gutenberg Project, 2707 works by 316 authors born in the 19th century, with at least 3 works from each author. ANOSIM (Analysis of similarities) [43] is used to test whether the inter-author differences are significantly greater than intra-author differences.

As can be seen in Fig. 3, ANOSIM indicated that intra-author language consistency outstrips inter-author consistency: works of different authors present greater differences in word frequency distribution than works of the same author [$R = 0.9346$ and $p = 0.0010$]. In other words, individual deviations from overall word frequency distribution may indicate stylistic differences among authors. This finding suggests that word frequency distributions extracted from language use are probably important indicators of personal stylistic features, and thus cannot be overlooked in stylistics and social linguistics.

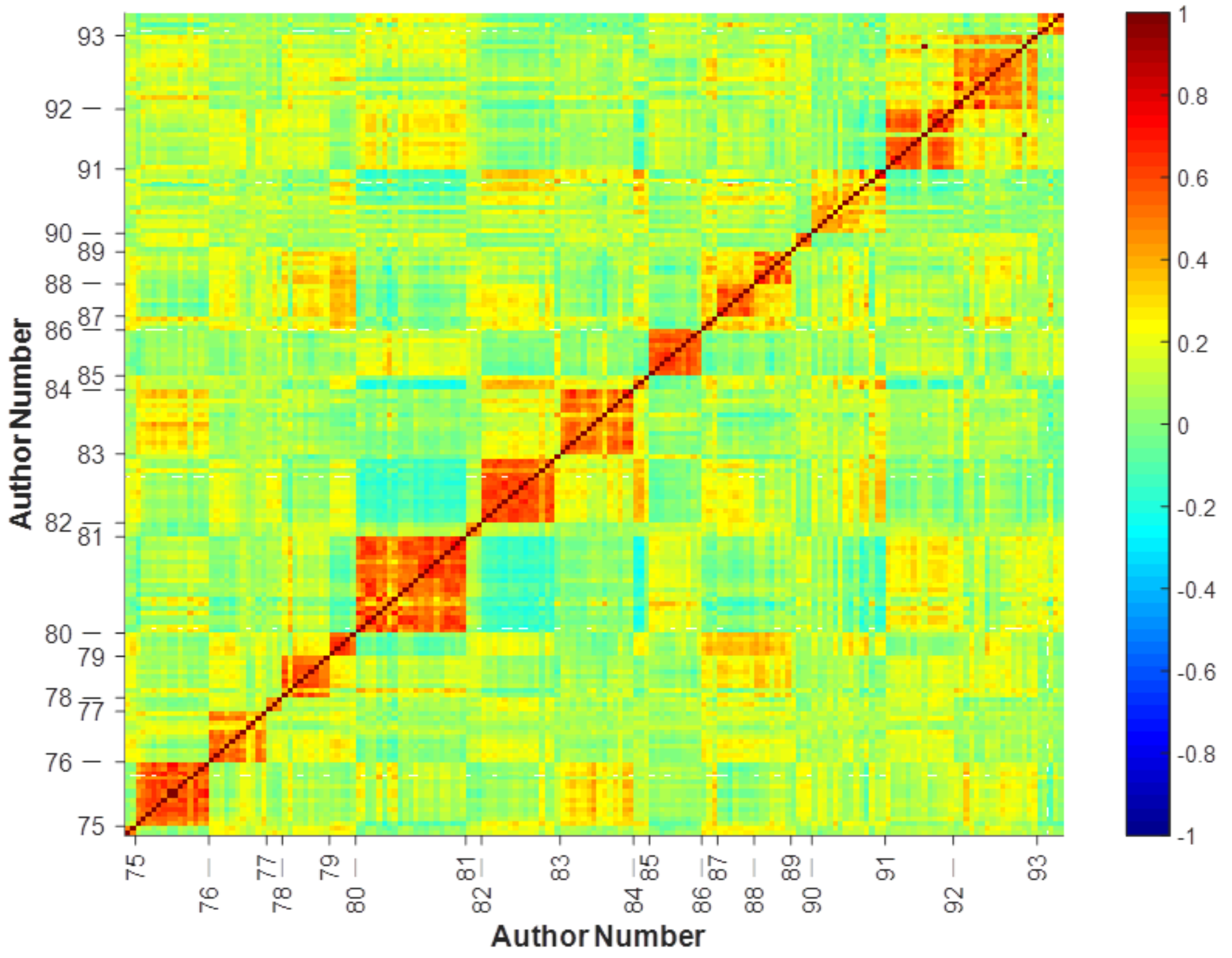

Fig. 3. Intra-author language consistency outstrips inter-author consistency. In the figure is a correlation coefficients matrix of 180 works. The works of the same author are arranged consecutively on X and Y axes, forming intervals marked by numbers representing those authors. The details are available in supplementary materials. In this figure Red points represent high correlation coefficients, which are largely distributed in the squares along the diagonal line of the matrix between authors, that is, the coefficients are much higher among the works of the same author. Here, we choose not to present all 2707 works because that will overcrowd the figure, rendering it unreadable.

## 4. Discussion and Conclusion

This study shows that word frequencies extracted from language use have some universal statistical patterns, and can capture language variations like diachronic drifts and personal styles.

The languages investigated all present power-law distributions, with the curves universally composed of two parts: the upper one of high-frequency words and the lower one of low-frequency words. In fact, the upper part universally presents two segments respectively composed of function words and basic-category notional words. This pattern of word frequency distribution is probably a fundamental linguistic universal, shaped by the dual-process mechanism of human cognition.

Dual-process model characterizes most high-level human cognitive activities: type-1 process features promptness, automaticity, effortlessness, and freedom from conscious attention, in most cases due to frequent repetition (high frequency); type-2 process, in contrast, features slowness, controlled attention, conscious effort, etc. [44].

The universal pattern of word frequency distribution may have much to do with Type-1 process, which automatizes language comprehension and organization to a considerable degree. The limited number of function words or basic category notional words are highly frequent, and high frequency render them readily accessible, subject to Type-1 processing. These words may form some fundamental templates in language processing or represents some most important blocks of our knowledge that deserve automatic processing. Such a pattern of word frequency distribution is perhaps molded by the common human cognition mechanism under the restriction of principle of least effort: a limited number vital to language are processed automatically owing to their high frequency. Thus, this distribution of word frequency is a linguistic universal, shaped by fundamental common cognitive mechanism and reflecting overall tendency in language use. But it is not a rigid yes/no rule, flexibly allowing for deviations and exceptions at microscopic scales. In the past, such exceptions have afflicted many researchers of language universals, with Greenberg probably as an exception, who introduced probability into language universals based on authentic language data [45].

Diachronic drifts are the key concerns of historical linguistics. Our findings suggest that word frequency distributions may quantitatively present the diachronic drifts of language system. This drift is gradual [46], which may be reflected by the slow change in the kernel words and their frequency distributions. Hence, the diachronic drift of language is a gradual change in the use of language, or rather in the frequency of language use. In the present study, such variations are quantitatively captured and described by the frequency distribution of kernel words. We may well believe that the diachronic drift can also be captured by frequency of other linguistic units or relations. No diachronic change in language is sudden and abrupt: it is always embodied in the change of frequency, or, the probability of use. The change is at first unnoticeably slight, but the slight change may implicitly influence our knowledge of the language, leading to change in the probability of use, which, under some conditions, may accumulate, multiply, cause chain reaction, and eventually become much drastic and significant over time. In most cases, rigid rules cannot satisfactorily explain such changes, which can only be found in terms of frequency and probability.

Similarly, the differences among social varieties of language may largely be the differences in frequency and probability, which is also a matter of degree. According to Saussure, langue is an idealized "mean" of the parole in a society, a "mean" replicated in individuals as slightly different copies, owing to different linguistic environments or different verbal habits. These linguistic environments and the verbal habits will decide how frequently or how probably a person will use a linguistic unit or structure, and different environment will lead to different habits, different frequencies or probabilities in language use, which make slight deviations from the "mean". To recap, these slight deviations of individual copies are mostly likely to be differences in frequency of some units or structures, not the differences in those units per se. Otherwise it is impossible for inter-personal communication if person use different linguistic units. So, personal styles probably consist in, to a considerable degree, the differences in frequency and probability of words used by individuals. This is supported by the present study, which indicates that differences in personal styles consist at least partly in the variations in frequency distribution. Such differences are usually matter of degree, beyond the reach of the rules of formal linguistic theories.

The present study confines itself to frequency of words, not involving the frequency or probability of grammatical constructions of word combinations, and thus having little to do with the so called grammaticality. But it is highly probable that the statistical patterns at the syntactic level are also important properties of languages, closely bearing on our judgment of grammaticality, and underlying our production and comprehension of sentences [47]. Anyway, grammaticality is a matter of degree, a matter of probability, somehow based on the frequency of words in language use.

There are increasing studies suggesting that language is a probabilistic system driven by human users [31]. The present one, based on large-scale multiple language data, provides for the first time macroscopic and systemic evidences that human languages are probably probabilistic systems. Languages are by nature probabilistic, and any linguistic theory neglecting statistical patterns in actual use will be divorced from the reality of language, failing to faithfully model the real language, to satisfactorily explain the language processing mechanism of human beings, and eventual to effectively unveil the mysteries of human intelligence and knowledge. To model language solely in terms of rules of dichotomy probably neglects many possibilities between YES and NO, providing only an over-simplified sketch of human language. In comparison, the conception that language is a probabilistic system may provide a picture of language with higher resolution. To faithfully model the real language, linguistic theories should pay enough attention to the probabilistic nature of language, and incorporate statistical patterns of human languages. Such theories may provide reliable guide for practical applications like NLP, propel the scientific explorations in linguistics, and more importantly, help unveil the mysteries of human languages and human intelligence.

(2018-10-30)